\documentclass[runningheads]{llncs}
\usepackage{graphicx}
\usepackage{comment}
\usepackage{amsmath,amssymb} 
\usepackage{color}
\usepackage{verbatim}

\usepackage{array}
\newcommand{\PreserveBackslash}[1]{\let\temp=\\#1\let\\=\temp}
\newcolumntype{C}[1]{>{\PreserveBackslash\centering}p{#1}}
\newcolumntype{R}[1]{>{\PreserveBackslash\raggedleft}p{#1}}
\newcolumntype{L}[1]{>{\PreserveBackslash\raggedright}p{#1}}

\usepackage{amssymb}

\usepackage[pagebackref=true,breaklinks=true,linkcolor=red,anchorcolor=blue,            citecolor=green,letterpaper=true,colorlinks,bookmarks=false]{hyperref}

\begin{document}
\pagestyle{headings}
\mainmatter
\def\ECCVSubNumber{4867}  

\title{Dynamic Group Convolution for Accelerating Convolutional Neural Networks}

\titlerunning{Dynamic Group Convolution for Accelerating CNNs}
%
\author{Zhuo Su\inst{1,\star} \and
Linpu Fang\inst{2,}\thanks{Equal contributions. $\dagger$ Corresponding author: li.liu@oulu.fi} \and
Wenxiong Kang\inst{2} \and \\ 
Dewen Hu\inst{3} \and
Matti Pietik{\"a}inen\inst{1} \and
Li Liu\inst{3,1,\dagger}
}
\authorrunning{Z. Su, L. Fang et al.}
%
\institute{Center for Machine Vision and Signal Analysis, University of Oulu, Finland \and
South China University of Technology, China \and
National University of Defense Technology, China
}
\maketitle


\begin{abstract}
Replacing normal convolutions with group convolutions can significantly increase the computational efficiency of modern deep convolutional networks, which has been widely adopted in compact
network architecture designs. However, existing group convolutions undermine the original network
structures by cutting off some connections permanently resulting in significant accuracy degradation. In
this paper, we propose dynamic group convolution (DGC) that adaptively selects which part of input channels  to be connected within each group for individual samples on the fly. Specifically, we equip each group with a small feature selector to automatically select
the most important input channels conditioned on the input images. Multiple groups can adaptively capture abundant and complementary visual/semantic features for each input image. The DGC preserves the original network structure and has similar computational efficiency as the conventional group convolution simultaneously. Extensive experiments on multiple image classification benchmarks including CIFAR-10, CIFAR-100 and ImageNet demonstrate its superiority over the existing group convolution techniques and dynamic execution methods\footnote{This work was partially supported by the Academy of Finland  and the National Natural Science Foundation of China under Grant 61872379.
The authors also wish to acknowledge CSC IT Center for Science, Finland, for computational resources.
}. The code is available at \url{https://github.com/zhuogege1943/dgc}.
\end{abstract}

\section{Introduction}
\label{section:introduction}
\begin{figure}[ht]
    \includegraphics[width=12cm]{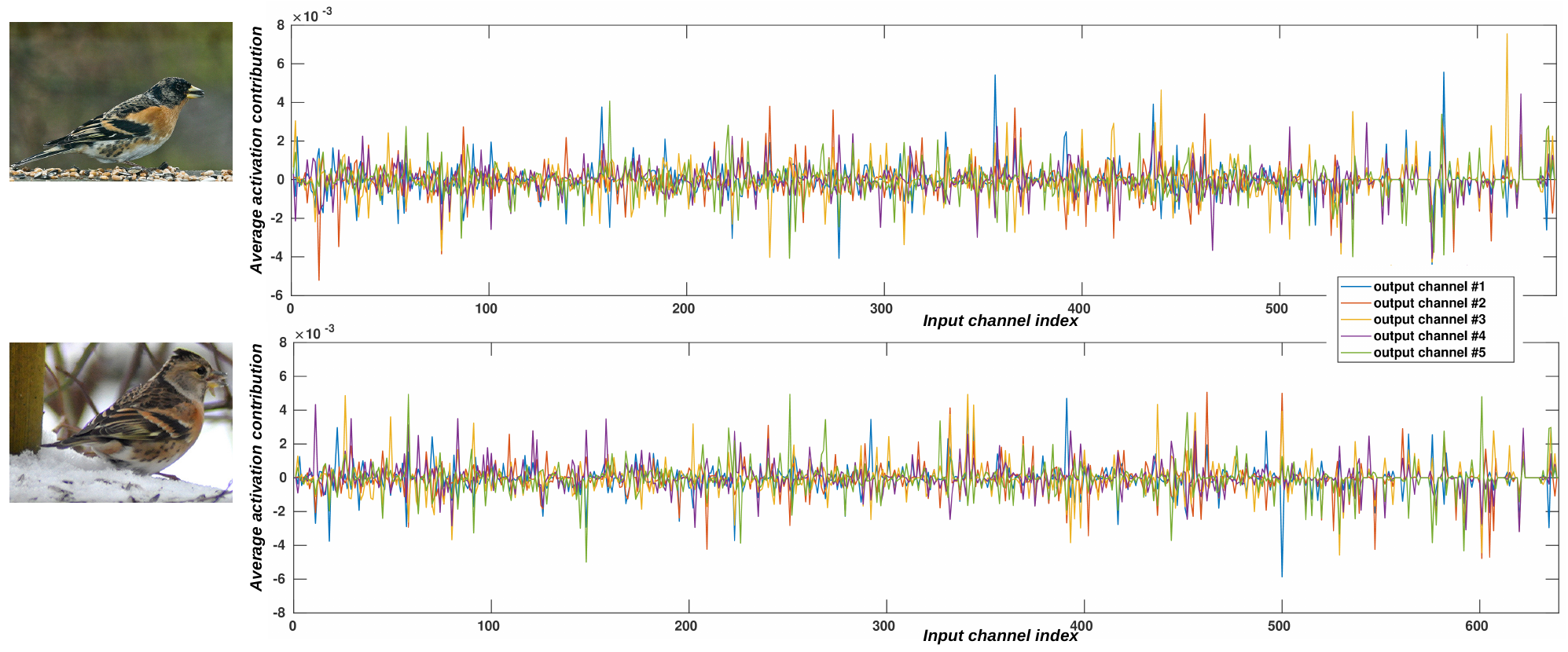}
    \vspace{-0.1in}
    \caption{The average contributions from the input channels to several output channels at a certain layer in a trained DenseNet~\cite{huang2017densely} are depicted. The X-axis and Y-axis indicate the input channels and their average contributions to the output channels. For a pair of input and output channel, an activation map is firstly obtained by convolution with the corresponding filter, and the average contribution from this input channel to the output channel is calculated as the average value of the activation map. It can be seen that such contributions vary a lot across different input-output pairs, which is the main observation behind the motivation of the proposed DGC.}
    \vspace{-0.2in}
    \label{fig:activations}
\end{figure}

Deep convolutional neural networks (CNNs) have achieved significant successes in a wide range of computer vision tasks including image classification \cite{deng2009imagenet}, object detection \cite{lin2014microsoft}, and semantic segmentation \cite{long2015fully}. Earlier studies \cite{huang2017densely,he2016deep} found that deeper and wider networks could obtain better performance, which results in a large number of huge and complex models being designed in the community. However, these models are very compute-intensive making them impossible to be deployed on edge devices with strict latency requirements and limited computing resources. In recent years, more and more researchers turn to study network compression techniques or design computation-efficient architectures to solve this troublesome.
\par Group convolution, which was first introduced in AlexNet \cite{krizhevsky2012imagenet} for accelerating the training process across two GPUs, has been comprehensively applied in computation-efficient network architecture designs \cite{zhang2018shufflenet,ma2018shufflenet,xie2017aggregated,chollet2017xception,sun2018igcv3}. Standard group convolution equally split the input and output channels in a convolution layer into \textit{G} mutually exclusive groups while performing normal convolution operation within individual groups, which reduces the computation burden by \textit{G} times in theory. The predefined group partition in standard group convolution may be suboptimal, recent studies \cite{wang2019fully,huang2018condensenet} further propose learnable group convolution that learns the connections between input and output features in each group during the training process.
\par By analyzing the existing group convolutions, it can be found that they have two key disadvantages: 1) They weaken the representation capability of the normal convolution by introducing sparse neuron connections and suffer from decreasing performance especially for those difficult samples; 2) They have fixed neuron connection routines, regardless of the specific properties of individual inputs. However, the dependencies among input and output channels are not fixed and vary with different input images, which can be observed in Fig. \ref{fig:activations}. Here, for two different input images, the average contributions from the input channels to several output channels at a certain layer in a trained DenseNet \cite{huang2017densely} are depicted. Two interesting phenomenons can be found in Fig. \ref{fig:activations}. Firstly, an output channel may receive information from input channels with varying contributions depending on a certain image, some of them are negligible. Secondly, for a single input image, some input channels with such negligible contributions correspond to groups of output channels, \textit{i.e.}, the corresponding connections could be cut off without influences on the final results. It indicates that it needs an adaptive selection mechanism to select which set of input channels to be connected with output channels in individual groups.
\par Motivated by the dynamical computation mechanism in dynamic networks \cite{lin2017runtime,gao2018dynamic,hua2019channel}, in this paper, we propose dynamic group convolution (DGC) to adaptively select the most related input channels for each group while keeping the full structure of the original networks. Specifically, we introduce a tiny  auxiliary feature selector for each group to dynamically decide which part of input channels to be connected based on the activations of all of input channels. Multiple groups can capture different complementary visual/semantic features of input images, making the DGC powerful to learn plentiful feature representations. Note that the computation overhead added by the auxiliary feature selectors are negligible compared with the speed-up provided by the sparse group convolution operations. In addition, the proposed DGC is compatible with various exiting deep CNNs and can be easily optimized in an end-to-end fashion.
\par We embed the DGC into popular deep CNN models including ResNet \cite{he2016deep}, CondenseNet \cite{huang2018condensenet} and MobileNetV2 \cite{sandler2018mobilenetv2}, and evaluate its effectiveness on three common image recognition benchmarks: CIFAR-10, CIFAR-100 and ImageNet. The experimental results indicate the DGC outperforms the exiting group convolution techniques and dynamic execution methods.

\section{Related Work}

\textbf{Efficient Architecture Design.} As special cases of sparsely connected convolution, group convolution and its extreme version, \textit{i.e.} depth-wise separable convolution, are most popular modules employed in efficient architecture designs. AlexNet \cite{krizhevsky2012imagenet} firstly uses group convolution to handle the problem of memory limitation. 
ResNeXt \cite{xie2017aggregated} further applies group convolution to implement a set of transformations and demonstrates its effectiveness. A series of subsequent researches use group convolution or depth-wise separable convolution to designe computation-efficient CNNs \cite{howard2017mobilenets,sandler2018mobilenetv2,zhang2018shufflenet,ma2018shufflenet,zhang2017interleaved,sun2018igcv3}. Instead of predefining the connection models, CondenseNet \cite{huang2018condensenet} and FLGC \cite{wang2019fully} propose to automatically learn the connections of group convolution during the training process. All these exiting group convolutions have fixed connections during inference, inevitably weakening the representation capability of the original normal convolutions due to the sparse structures. Our proposed DGC can effectively solve this troublesome by employing dynamic execution strategy that keeps sparse computation without undermining original network structures.  

\noindent\textbf{Network Compression.} Generally, compression methods can be categorized into five types: quantization \cite{cao2019seernet,liu2019circulant,ding2018universal,sakr2018per,banner2019post}, knowledge distillation \cite{heo2019comprehensive,jin2019knowledge,phuong2019towards,yoo2019knowledge,liu2019knowledge}, low-rank decomposition \cite{minnehan2019cascaded,denton2014exploiting,jaderberg2014speeding,zhang2015efficient}, weight sparsification \cite{han2015learning,lee2018snip,xiao2019autoprune}, and filter pruning \cite{liu2019variational,li2019exploiting,he2019filter,peng2019collaborative}.  Quantization methods accelerate deep CNNs by replacing high-precision float point operations with low-precision fixed point ones, which usually incurs significantly accuracy drop. Knowledge distillation methods aim to learn a small student model by mimicking the output or feature distributions of a larger teacher model. Low-rank decomposition methods reduce the computation by factorizing the convolution kernels. Weight sparsification methods removes individual connections between neural nodes resulting in irregular models, while filter pruning methods, which our method is most related to, directly cut off entire filters keeping the regular architectures. Generally, those compression methods usually need several rounds to obtain the final compact models. On the contrary, the proposed DGC can be easily optimized with any exiting networks in an end-to-end manner. Note that the filter pruning removes some channels after training, which loses the capabilities of the original CNNs permanently, while our method keeps the capabilities by the dynamic channel selection scheme.  

\noindent\textbf{Dynamic Network.} In contrast to the one-size-fit-all paradigm, dynamic networks \cite{wang2018skipnet,liang2019learning,wu2018blockdrop,liu2018dynamic,phuong2019distillation,huang2017multi,dong2017more} dynamically execute different modules across a deep network for individual inputs. One kind of method \cite{bolukbasi2017adaptive,phuong2019distillation,huang2017multi} sets multiple classifiers at different places in a deep network and apply early exiting strategy to implement dynamic inference. Another line of work \cite{wang2018skipnet,liang2019learning,veit2018convolutional,lin2017runtime,wu2018blockdrop,liu2018dynamic,dong2017more} learns auxiliary controllers or gating units to decide which parts of layers or channels could be skipped for extracting intermediate feature representations. Among those dynamic execution methods, our proposed DGC is mostly motivated
by the channel or pixel-level ones \cite{lin2017runtime,gao2018dynamic,hua2019channel}. Runtime Neural Pruning (RNP) \cite{lin2017runtime} learns a RNN to generate binary policies and adaptively prune channels in convolutional layers. The RNN is trained by reinforcement learning with a policy function, which is hard to converge and sensitive to hyper-parameters. Feature boosting and suppression (FBS) \cite{gao2018dynamic} generates decisions to dynamically prune a subset of output channels, which is prone to suffer from the \textit{internal covariate shift} problem~\cite{ioffe2015bn} due to its interference in batch normalization (BN), bringing instability to training.
Channel gating neural network (CGNet) \cite{hua2019channel} uses a subset of input channels to generate a binary decision map, and skips unimportant computation in the rest of input channels. The CGNet actually works in a semi-dynamic way and uses a complex approximating non-differentiable gate function to learn the binary decisions. In contrast to these methods, our proposed DGC works better to achieve stable training with dynamic channel selection, and can be easily optimized using common SGD algorithms.  

\section{Dynamic Group Convolution}
\begin{figure}[ht]
    \includegraphics[width=12cm]{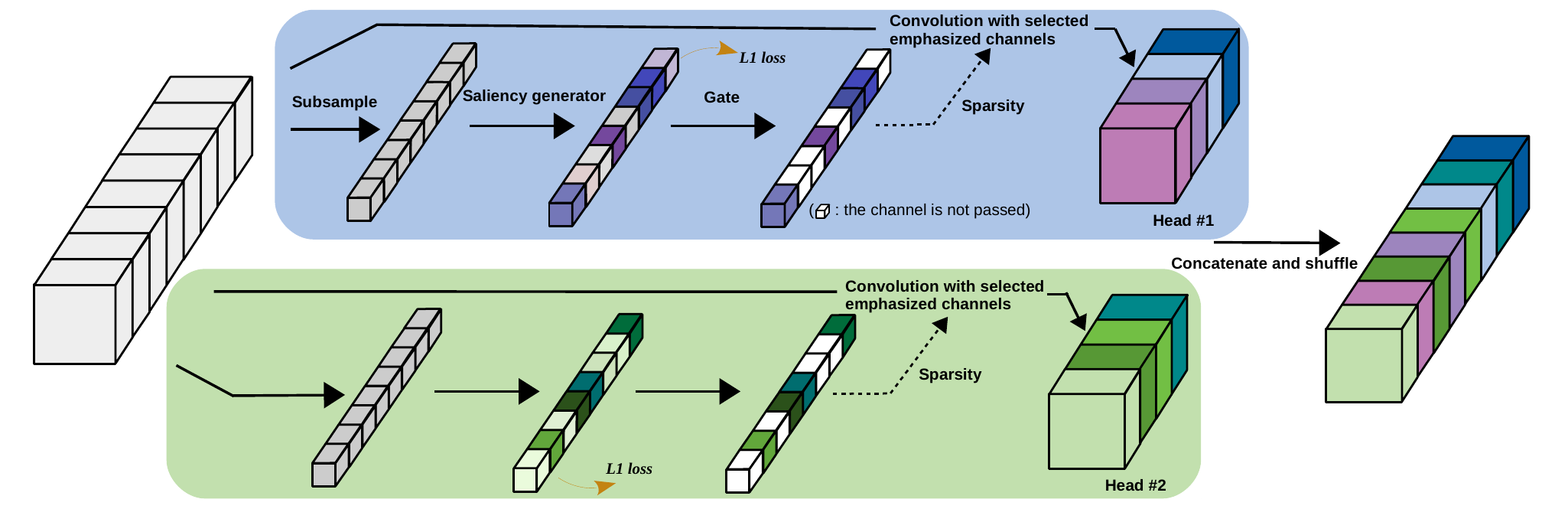}
    \vspace{-0.1in}
    \caption{Overview of a DGC layer. For a simple illustration, we set the number of channels as 8 for both input and output volume and the number of heads as 2. Each head is responsible for generating half of the output volume, with different part of the input volume, by respectively using a saliency guided channel selector (gate). Here, white blocks after the gate represents the corresponding channel would not take part in the convolution.}
    \vspace{-0.15in}
    \label{fig:structure}
\end{figure}
\subsection{Group Convolution and Dynamic Channel Pruning Revisiting}
\label{section:revisiting}
Consider how a CNN layer works with groups, which can be generalized as:
{\small \begin{align}
    & \pmb{x}' = [\hat{f}(\pmb{x}^{1}), \hat{f}(\pmb{x}^{2}), ..., \hat{f}(\pmb{x}^{N})]\label{eq:1}, \\
    & \hat{f}(\pmb{x}^{i}) = f(\pmb{x}^{i}, \pmb{\theta}^{i})\cdot \pi(\pmb{x}^{i},\pmb{\phi}^{i})\label{eq:2},
\end{align}}
where $\pmb{x} \in \mathbb{R}^{C \times H \times W}$, $\pmb{x}'\in \mathbb{R}^{C'\times H' \times W'}$ are the input and output of the current layer respectively, $\pmb{x}^{i}\subseteq{\pmb{x}}$ is a subset of input channels that is divided into the \textit{ith} group, $N$ is the number of groups, $f$ executes the conventional convolution using parameter $\pmb{\theta}$, while the meaning of $\pi$ is case-dependent, with its parameter $\pmb{\phi}$, which would be discussed below. 

In a standard convolution layer, $N=1$ and $\pi \equiv \pmb{1}$. While a standard group convolution (SGC) evenly allocates equal number of channels in $\pmb{x}$ to each group with hard assignments, thus reducing the computation cost by $N$ times. On the other hand, studies show that although complemented with channel permutation before performing group convolution, SGC is not always optimal \cite{huang2018condensenet,wang2019fully}, and introduce a learning strategy to soften the hard assignments, leading to a better group allocation. Such soft assignments allow each channel in $\pmb{x}$ to automatically choose its fitted group(s). To achieve the soft assignments, common methods are to add a group-lasso loss to induce group-level weights sparsity or a learnable binary decision matrix \cite{huang2018condensenet,wang2019fully}. However, these methods can not be easily transferred to dynamic versions due to their ``staticness''.

In a convolution layer with dynamic execution \cite{gao2018dynamic,dong2017more,hua2019channel,lin2017runtime} and $N=1$, a on-the-fly gate score or saliency is pre-calculated through a prediction function (represented as $\pi$ in \eqref{eq:2}) for each output cell that needs to be computed by $f$. The cell could be an output channel~\cite{gao2018dynamic,lin2017runtime} or a pixel in the output feature maps~\cite{dong2017more,hua2019channel}. Computation of the output cell in $f$ is skipped if the corresponding saliency is small enough.
Denoting the saliency vector (output of $\pi$, with size equal to the number of output cells) as $\pmb{g}$, sparsity for $\pmb{g}$ is implemented by reinforcement learning in~\cite{lin2017runtime}, which is hard to be optimized. \cite{dong2017more,gao2018dynamic} associates $\pmb{g}$ with a lasso loss, which is in fact a L1 norm regularization:
{\small \begin{equation}
    \label{eq:3-lasso}
    L_{lasso} = \left\Vert g_{1}, g_{2}, ...\right\Vert_{1}.
\end{equation}}
In \cite{dong2017more}, $\pi$ and $f$ are homogeneous functions (both are convolutions and compute tensors with the same shape), halving $\pmb{\phi}$ and doubling $\pmb{\theta}$ can always lead to decrease in \eqref{eq:3-lasso}, making the lasso loss meaningless. \cite{gao2018dynamic} regards $\pmb{g}$, where $\pmb{g}\in \mathbb{R}^{1\times C'}$, as a replacement of the BN scaling factors, so the following BN layer is modified by eliminating the scaling parameters. In such case, function $\pi$ need to be carefully designed and trained due to the \textit{internal covariate shift} problem~\cite{ioffe2015bn} (will be analyzed in the next section). 

Different from the above mentioned dynamic execution methods that skips computations for some output cells, the dynamic mechanism in DGC conducts $\pmb{g}$ on the input channels to sparsify their connections with the output channels, while keeping the shape of the output volume unchanged.
Embedding into group executions, DGC actually performs like the multi-head self-attention mechanism proposed in \cite{vaswani2017attention}. Thereby, the Eq.~\ref{eq:2} is modified as:
{\small \begin{equation}
    \label{eq:4-modfunction}
    \hat{f}(\pmb{x}^{i}) = f(\pmb{x}^{i}\cdot\pi(\pmb{x}^{i},\pmb{\phi}^{i}), \pmb{\theta}^{i}).
\end{equation}}
In this case, the scale-sensitive problem can be easily removed by connecting an untouched BN layer (see Section~\ref{section: GDC}). 
Note that in practical computation, $\pmb{\theta}^{i}$ is also partially selected according to the selected channels in $\pmb{x}^{i}$ (see Eq.~\ref{eq:emphasized-conv}).


\subsection{Group-wise Dynamic Execution}
\label{section: GDC}
An illustration of the framework of a DGC layer can be seen in Fig.~\ref{fig:structure}. We split the output channels into multiple groups, each of them is generated by an auxiliary head that equips with an input channel selector to decide which part of input channels should be selected for convolution calculation (see the blue and green areas in Fig.~\ref{fig:structure}). First, the input channel selector in each head adopts a gating strategy to dynamically determine the most important subset of input channels according to their importance scores generated by a saliency generator. Then, the normal convolution is conducted based on the selected subset of input channels generating the output channels in each head. Finally, the output channels from different heads are concatenated and shuffled, which would be connected to a BN layer and non-linear activation layer.

\medskip
\noindent{\bf Saliency Generator.} The saliency generator assigns each input channel a score representing its importance. Each head has a specific saliency generator, which encourages different heads to use different subpart of the input channels and achieve diversified feature representations. In this paper, we follow the design of the SE block in \cite{hu2018squeeze} to design the saliency generator. For the \textit{ith} head, the saliency vector $\pmb{g}^{i}$ is calculated as:
{\small \begin{equation}
    \label{eq:5}
    \pmb{g}^{i} = \pi(\pmb{x}^{i},\pmb{\phi}^{i}) = \pi(\pmb{x},\pmb{\phi}^{i}) = (W^{i}(p(\pmb{x})) + \pmb{\beta}^{i})_{+},
\end{equation}}
where $\pmb{g}^{i}\in\mathbb{R}^{1\times C}$ represents the saliency vector for the input channels, $(z)_+$ denotes the ReLU activation, $p$ reduces each feature map in $\pmb{x}$ into a single scalar, such as global average pooling as we used in our experiments, $\pmb{\beta}^{i}$ and $W^{i}$ are trainable parameters representing the biases and a two-step transformation function mapping $\mathbb{R}^{1\times C}\mapsto\mathbb{R}^{1\times C/d}\mapsto\mathbb{R}^{1\times C}$ with $d$ being the squeezing rate. Note that $\pmb{x}^{i}$ in eq.~\ref{eq:4-modfunction} is equal to the whole input volume $\pmb{x}$ here, meaning all the input channels would be considered as candidates in each of the heads.

\medskip
\noindent{\bf Gating Strategy.}
Once the saliency vector is obtained, the next step is to determine which part of the input channels should join the following convolution in the current head.
We can either use a head-wise or globally network-wise threshold that decides a certain number of passed gates out of all for each head in a DGC layer. We adopt the head-wise threshold here for simplicity and the global threshold will be discussed in \ref{section: imagenet}.

Given a target pruning rate $\xi$, the head-wise threshold $\tau^{i}$ in $ith$ head meets the equation where $|\mathcal{S}|$ means the length of set $\mathcal{S}$:
{\small \begin{equation}
\label{eq:threshold}
    \xi = \frac{|\{g^{i}\; |\; g^{i}<\tau^{i}, g^{i}\in \pmb{g}^{i}\}|}{|\{g^{i}\; |\; g^{i}\in \pmb{g}^{i}\}|}.
\end{equation}}
The saliency goes with two streams, \textit{i.e.}, any of channels in the input end with its saliency smaller than the threshold is screened out, while anyone in the remaining channels is amplified with its corresponding $g$, leading to a group of selected emphasized channels $\pmb{y}^i\in\mathbb{R}^{(1-\xi)C\times H\times W}$. Assuming the number of heads is $\mathcal{H}$, the convolution in the \textit{ith} head is conducted with $\pmb{y}^i$ and the corresponding filters $\pmb{w}^{i}$ ($\pmb{w}^{i}\subset\pmb{\theta}^{i}$, $\pmb{\theta}^{i}\in\mathbb{R}^{k\times k\times C\times \frac{C'}{\mathcal{H}}}$ and $k$ is the kernel size):
{\small \begin{align}
    \hat{\pmb{x}}^{i} = \pmb{x}[\pmb{\upsilon},:,:],\:\:\: \hat{\pmb{g}}^{i} = \pmb{g}^{i}[\pmb{\upsilon}],\:\:\: \pmb{w}^{i} =\pmb{\theta}^i[:,:,\pmb{\upsilon},:],\:\:\: \pmb{\upsilon} = \mathcal{I}_{top}\lceil(1-\xi)C\rceil(\pmb{g}^{i}) \nonumber\\
    \pmb{x}'^{i} = \hat{f}(\pmb{x}^{i}) = f(\underbrace{\pmb{x}\cdot\pi(\pmb{x},\pmb{\phi}^{i})}_{\textrm{before pruning}}, \pmb{\theta}^{i}) = f(\underbrace{\hat{\pmb{x}}^{i}\cdot\hat{\pmb{g}}^{i}}_{\textrm{after pruning}}, \pmb{w}^{i}) = f(\pmb{y}^{i}, \pmb{w}^{i}) = \pmb{y}^{i}\otimes\pmb{w}^{i}, \label{eq:emphasized-conv}
\end{align}}
where $\mathcal{I}_{top}\lceil \mathrm{k}\rceil(\pmb{z})$ returns the indices of the $\mathrm{k}$ largest elements in $\pmb{z}$, 
the output $\pmb{x}'^{i}\in\mathbb{R}^{\frac{C'}{\mathcal{H}}\times H'\times W'}$, $\otimes$ means the regular convolution.
At the end of the DGC layer, individual outputs gathered from multiple heads are simply concatenated and shuffled, producing $\pmb{x}'$.
The gating strategy is unbiased for weights updating since all the channels own the equal chance to be selected.

To induce sparisty, we also add a lasso loss following Eq.~\ref{eq:3-lasso} as
{\small \begin{equation}
    \label{eq:dgc-lasso}
    L_{lasso} = \lambda\frac{1}{\mathcal{L}\mathcal{H}}\sum_{l=1}^{\mathcal{L}}\sum_{i=1}^{\mathcal{H}}\left\Vert\pmb{g}^{l,i}\right\Vert_{1}
\end{equation}}
to the total loss, where $\mathcal{L}$ is the number of DGC layers and $\lambda$ is a pre-defined parameter.

\medskip
\noindent{\bf Computation Cost.}
Regular convolution using the weight tensor $\pmb{\theta}$ with kernel size $k$ takes $k^2C'CH'W'$ \textit{multiply-accumulate operations} (MACs). In a DGC layer for each head, 
the saliency generator and convolution part costs {\small $\frac{2C^2}{d}$} and {\small $k^2(1-\xi)C\frac{C'}{\mathcal{H}}H'W'$} MACs respectively. Therefore, the overall MAC saving of a DGC layer is:
{\small \begin{equation*}
    (k^2CC'H'W')/(\mathcal{H}(k^2(1-\xi)C\frac{C'}{\mathcal{H}}H'W' + \frac{2C^2}{d})) = 1/((1-\xi) + \frac{2\mathcal{H}C}{dk^{2}C'H'W'}) \approx 1/(1-\xi).
\end{equation*}}
As a result, the number of heads $\mathcal{H}$ has negligible influence on the total computation cost.

\begin{figure}[t]
\centering
    \includegraphics[width=12cm]{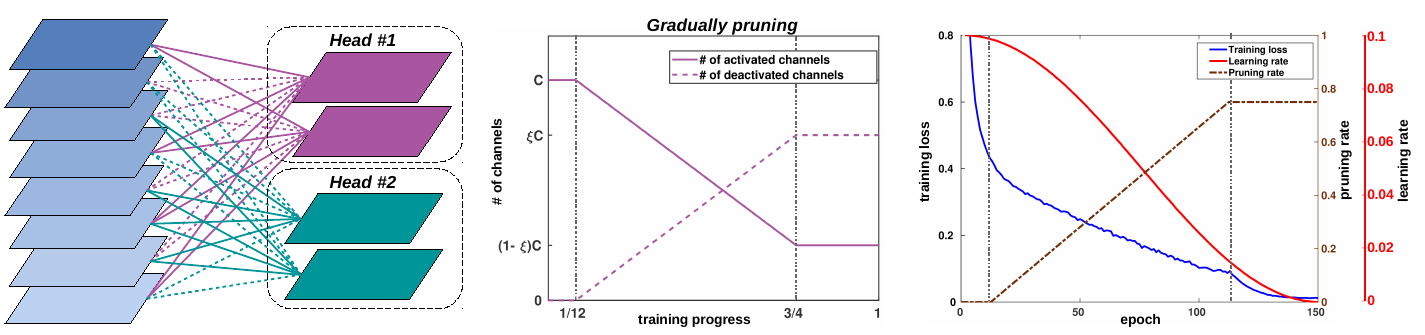}
    \vspace{-0.1in}
    \caption{The training process of a DGC network. We gradually increase the number of deactivated input channels in each DGC layer with three stages, using a cosine shape learning rate, to make the first stage (first $1/12$ of the training epochs, where no channels are pruned) warm up the network, the second stage perform gradual pruning, and the last stage (last $1/4$ epochs) fine-tune the sparsified structure. The right figure shows an example of training process on the CIFAR-10 dataset, with pruning rate 0.75 and total 150 epochs. The blue curve shows that even after entering the second stage, the training loss smoothly drops towards 0.}
    \vspace{-0.12in}
    \label{fig:training}
\end{figure}

\medskip
\noindent{\bf Invariant to Scaling.}
As discussed in Section~\ref{section:revisiting}, \cite{gao2018dynamic} implements a convolution layer (including normalization and non-linear activation) by replacing BN scaling factors with the saliency vector:
{\small \begin{equation*}
    \mathfrak{L}(\pmb{x}) = (\pmb{g}\cdot \mathsf{norm}(\mathsf{conv}(\pmb{x}, \pmb{\theta})) + \pmb{\beta})_{+},
\end{equation*}}
where $\mathsf{norm}(z)$ normalizes each output channel with mean $0$ and variance $1$. Since $\pmb{g} = \pi(\pmb{x},\pmb{\phi})$ is dynamically generated, scaling in $\pmb{\phi}$ first leads to scale change in $\pmb{g}$ and then in $\mathfrak{L}(\pmb{x})$, thus each sample has a self-dependent distribution, leading to the \textit{internal covariate shift} problem during training on the whole dataset \cite{ioffe2015bn}. In contrast, the inference with a DGC layer is defined as (we denote $\pmb{y}^{i}$ as $\pmb{y}$ and $\pmb{w}^{i}$ as $\pmb{w}$ for convenience):
{\small \begin{equation*}
    \mathfrak{L}(\pmb{x}) = (\mathsf{batchnorm}(\mathsf{conv}(\pmb{y}, \pmb{w})))_{+}.
\end{equation*}}
Combining Eq.~\ref{eq:emphasized-conv} and above equation, scaling in $\pmb{\phi}$ leads to scale change in $\pmb{g}$ while stops at $\pmb{y}$. With the following batch normalization, the ill effects of the internal covariate shift can be effectively removed \cite{ioffe2015bn}. 


\medskip
\noindent{\bf Training DGC Networks.} We train our DGC network from scratch by an end-to-end manner, without the need of model pre-training. During backward propagation, for $\pmb{\theta}$ which accompanies a decision process, gradients are calculated only for weights connected to selected channels during the forward pass, and safely set as 0 for others thanks to the unbiased gating strategy. To avoid abrupt changes in training loss while pruning, we gradually deactivate input channels along the training process with a cosine shape learning rate (see Fig.~\ref{fig:training}).

\section{Experiments}
\label{section:experiments}

\begin{figure}[ht]
    \centering
    \includegraphics[width=11cm]{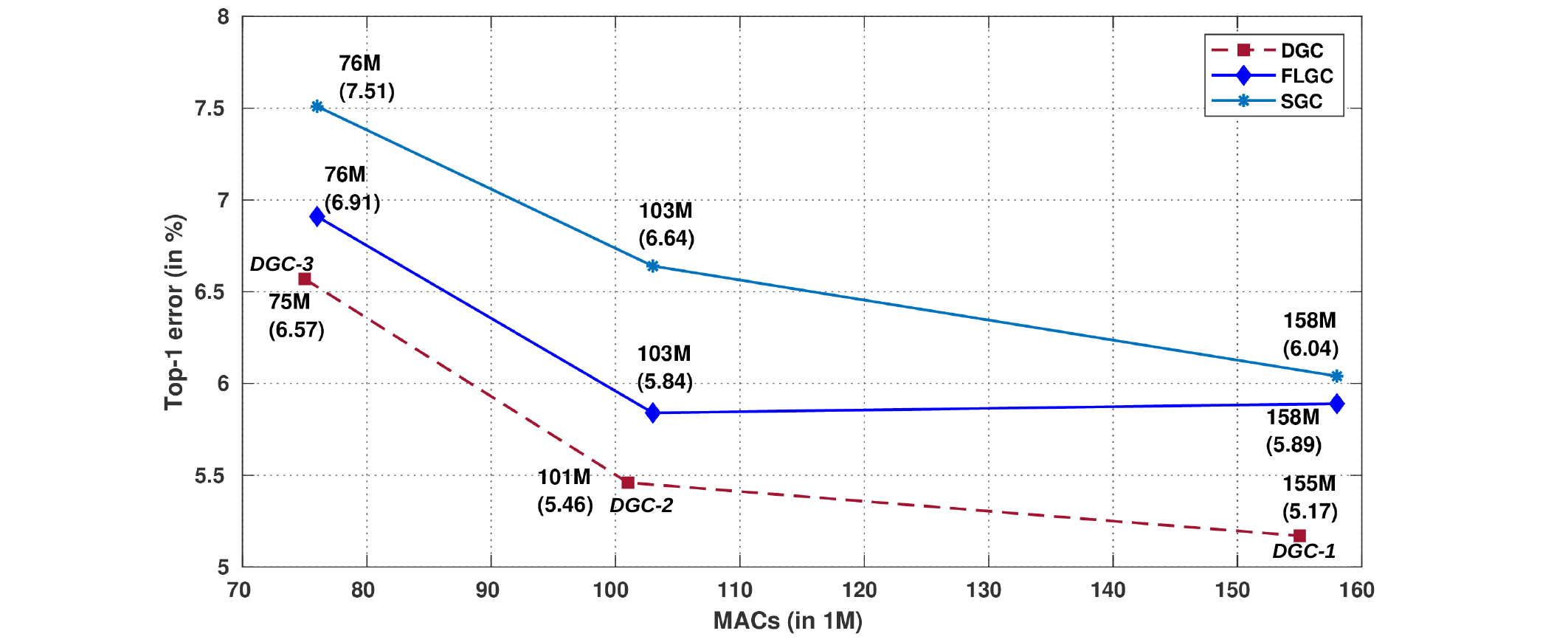}
    \vspace{-2mm}
    \caption{Comparison with SGC and FLGC based on MobileNetV2. The last 1$\times$1 convolution layers in all inverted residual blocks are replaced with corresponding group convolution layers. For DGC, the width multiplier and the positions of the downsample layers are adjusted to obtain models with matched MACs comparing with SGC and FLGC. For DGC-1 and DGC-2, the downsample layers are set on the first layer of the \textit{3rd} and \textit{5th} block. While the first layer of the \textit{2nd} and \textit{4th} block are set as the downsample layers for DGC-3. DGC-1 sets the width multiplier to be 1, while both DGC-2 and DGC-3 set the width multiplier to be 0.75. The results of SGC and FLGC are copied from the FLGC paper \cite{wang2019fully}.}
    \vspace{-0.2in}
    \label{fig:cifar_mobilenetv2}
\end{figure}

\subsection{Experimental settings}
We evaluate the proposed DGC on three common image classification benchmarks: CIFAR-10, CIFAR-100 \cite{krizhevsky2009learning}, and ImageNet (ILSVRC2012) \cite{deng2009imagenet}. The two CIFAR datasets both include 50K training images and 10K test images with 10 classes for CIFAR-10 and 100 classes for CIFAR-100. The ImageNet dataset consists of 1.28 million training images and 50K validation images from 1000 classes. For evaluations, we report detailed accuracy/MACs trade-off against different methods on both CIFAR and ImageNet dataset.

We choose previous popular models including ResNet \cite{he2016deep}, CondenseNet \cite{huang2018condensenet} and MobileNetV2 \cite{sandler2018mobilenetv2}as the baseline models and embed our proposed DGC into them for evaluations.

\smallskip
\noindent\textit{ResNet with DGC.} ResNet is one of the most powerful models that has been extensively applied in many visual tasks. In this paper, we use the ResNet18 as the baseline model and replace the two 3$\times$3 convolution layers in each residual block with the proposed DGC. 

\smallskip
\noindent\textit{MobileNetV2 with DGC.} MobileNetV2 \cite{sandler2018mobilenetv2} is a powerful efficient network architecture, which is designed for mobile applications. To further improve its efficiency, we replace the last 1$\times$1 point-wise convolution layer in each inverted residual block with our proposed DGC. 

\smallskip
\noindent\textit{CondenseNet with DGC.} CondenseNet is also a popular efficient model, which combines the dense connectivity in DenseNet \cite{huang2017densely} with a learned group convolution (LGC) module. We compare against the LGC by replacing it with the proposed DGC.  

\begin{figure}[ht]
\centering
    \includegraphics[width=12cm]{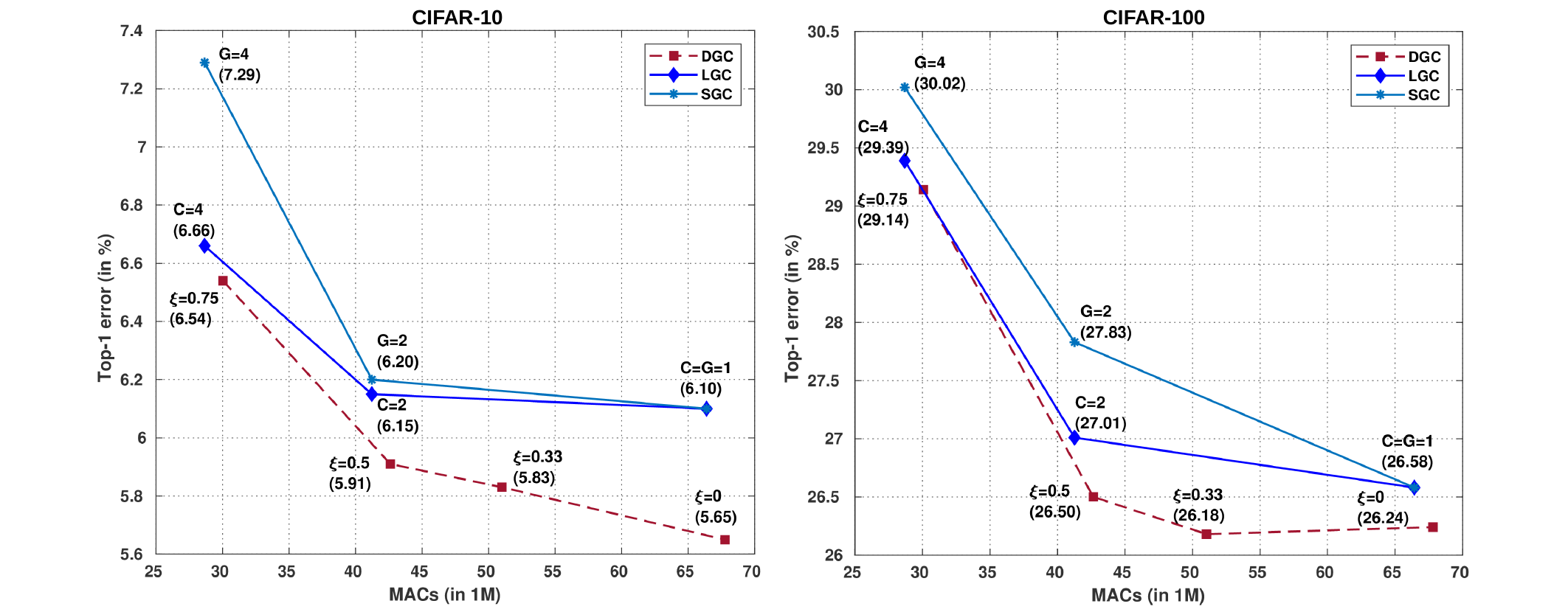}
    \vspace{-0.1in}
    \caption{Comparison with SGC and LGC based on CondenseNet on CIFAR-10 (left) and CIFAR-100 (right). The groups (heads) is set as 4 for both LGC and DGC. We change the pruning rate $\xi$ in DGC, \textit{condensation factor} $C$ in LGC and the number of groups $G$ in SGC to control the model computation costs. Noting that for LGC and SGC, when $C=G=1$, the two models actually share the same structure, corresponding to our DGC version when $\xi=0$. While a DGC layer works in a self-attention way with $\mathcal{H}$ auxiliary saliency generators, making it better than its vanilla counterparts.}
    \vspace{-0.15in}
    \label{fig:cifar_condensenet}
\end{figure}

All experiments are conducted using Pytorch \cite{paszke2019pytorch} deep learning library. 
For both datasets, the standard data augmentation scheme is adopted as in \cite{huang2018condensenet}. All models are optimized using stochastic gradient descent (SGD) with Nesterov momentum with a momentum weight of 0.9, and the weight decay is set to be $10^{-4}$. Mini batch size is set as 64 and 256 for CIFAR and ImageNet, repectively. The cosine shape learning rate (Fig.~\ref{fig:training}) starts from 0.1 for ResNet/CondenseNet and 0.05 for MobileNetV2. 
Models are trained for 300 epochs on CIFAR and 150 epochs on ImageNet for MobileNetV2, 150 epochs on CIFAR and 120 epochs on ImageNet for CondenseNet, and 120 epochs for ResNet on ImageNet. By default we use 4 groups (heads) in each DGC layer.

\subsection{Results on Cifar}

We compare DGC with both SGC and previous state-of-the-art learnable group convolutions including CondenseNet \cite{huang2018condensenet} and FLGC \cite{wang2019fully} on CIFAR datasets to demonstrate the effectiveness of its dynamic selecting mechanism. 
We use MobileNetV2 as backbone and conduct comparison with SGC and FLGC . For comparison with CondenseNet, 
we employ the proposed modified DenseNet structure in CondenseNet with 50 layers as the baseline model. 

\par \textbf{MobileNetV2.} We compare the proposed DGC with SGC and FLGC on CIFAR-10. Like the FLGC, we replace the last 1$\times$1 convolution layer in each inverted residual block with the proposed DGC. For SGC and FLGC, models with different MACs are obtained by adjusting the number of groups in group convolution layers. For our proposed DGC, we adjust the width multiplier and the positions of the downsample convolution layers (\textit{i.e.}, with stride equals to 2) to obtain models with matched MACs. The pruning rate $\xi$ is set to be 0.65 for each head and 4 heads are employed in a DGC layer. The results are shown in Fig.~\ref{fig:cifar_mobilenetv2}. It can be seen that both DGC and FLGC outperforms SGC, while our method can achieve lowest top-1 errors with less computation costs. Unlike the even partition of input channels and filters in SGC, FLGC adopts a grouping mechanism to dynamically determine the connections for each group during the training process, which can obtain more flexible and efficient grouping structures. The results in Fig.~\ref{fig:cifar_mobilenetv2} also demonstrate the effectiveness of FLGC over SGC. However, the group structures are still fixed in FLGC after training, which ignores the properties of single inputs and results in poor connections for some of them. By introducing dynamic feature selectors, our proposed DGC can adaptively select most related input channels for each group conditioned on the individual input images. The results indeed demonstrate the superiority of DGC compared with FLGC. Specifically, DGC achieves lower top-1 errors and computation cost (for example, 5.17\% \textit{vs.} 5.89\% and 155M \textit{vs.} 158M). 

\par \textbf{CondenseNet.} We compare the proposed DGC with SGC and CondenseNet on both CIFAR-10 and CIFAR-100. CondenseNet adopts a LGC strategy to sparsify the compute-intensive 1x1 convolution layers and uses the group-lasso regularizer to gradually prune less important connections during training. The final model after training is also fixed like FLGC. We replace all the LGC layers in CondenseNet with the SGC or DGC structures under similar computation costs for comparison. The results are shown in Fig.~\ref{fig:cifar_condensenet}. It also shows both DGC and LGC perform better than SGC due to the learnable soft channel assignments for groups. Meanwhile, DGC can perform better than LGC with less or similar MACs (for example, 5.83\% \textit{vs.} 6.10\% and 51M \textit{vs.} 66M on CIFAR-10, 26.18\% \textit{vs.} 26.58\% and 51M \textit{vs.} 66M on CIFAR-100), demonstrating that the dynamic computation scheme in DGC is superior to the static counterpart in LGC.  

\subsection{Results on ImageNet}
\label{section: imagenet}

\begin{table}[t]
\caption{Comparison of Top-1 and Top-5 classification error with state-of-the-art filter pruning and dynamic channel selection methods using ResNet-18 as the baseline model on ImageNet.}
\centering
\setlength{\tabcolsep}{8pt}
\resizebox*{10cm}{!}{
\begin{tabular}{lccccc}
\hline
\textbf{Model} & \textbf{Group} & \textbf{Dynamic} & \textbf{Top-1} & \textbf{Top-5} & \textbf{MAC Saving}\\
\hline
\hline
SFP \cite{he2018soft} & & & 32.90 & 12.22 & (1.72$\times$) \\
NS \cite{liu2017learning} & & & 32.79 & 12.61 & (1.39$\times$) \\
DCP \cite{zhuang2018discrimination} & & & 32.65 & 12.40 & (1.89$\times$) \\
FPGM \cite{he2019filter} & & & 31.59 & 11.52 & (1.53$\times$)\\
LCCN \cite{dong2017more} & & \checkmark & 33.67 & 13.06 & (1.53$\times$)\\
FBS \cite{gao2018dynamic} & & \checkmark & 31.83 & 11.78 & (1.98$\times$)\\
CGNet \cite{hua2019channel} & \checkmark & \checkmark & 31.70 & - & (2.03$\times$)\\
\textbf{DGC} & \textbf{\checkmark} & \textbf{\checkmark} & \textbf{31.22} & \textbf{11.38} & \textbf{(2.04$\times$)}\\
DGC-G & \textbf{\checkmark} & \textbf{\checkmark} & 31.37 & 11.56 & ($2.08\times$)\\
\hline
\end{tabular}
}
\vspace{-0.16in}
\label{table:ImageNet-1}
\end{table}
 
To further demonstrate the effectiveness of the proposed DGC, we compare it with state-of-the-art filter pruning and channel-level dynamic execution methods using ResNet18 as the baseline model on ImageNet. We also embed the DGC into the CondenseNet \cite{huang2018condensenet} and MobileNetV2 \cite{sandler2018mobilenetv2} (\textit{i.e.,} CondenseNet-DGC and MobileNetV2-DGC) to compare with state-of-the-art efficient CNNs. For CondenseNet-DGC/SGC, the same structure as CondenseNet in \cite{huang2018condensenet} is used with LGC replaced by DGC/SGC. For MobileNetV2-DGC, the last 1$\times$1 convolution layers in all inverted residual blocks in the original MobileNetV2 are replaced with DGC layers. We set $\mathcal{H}=4$, $\xi = 0.75$ and $\lambda = 10^{-5}$ for all models. The results are reported in Table \ref{table:ImageNet-1} and Table \ref{table:ImageNet-2}. 
Here, for further exploration, we additionally adopt a global threshold mentioned in \ref{section: GDC} to give more flexibility to the DGC network. The global threshold is obtained by firstly collecting all the saliency values throughout the whole network as $\pmb{G}$ and then replacing $\pmb{g}^{i}$ with $\pmb{G}$ in Eq.~\ref{eq:threshold}, with a third loss to reduce the inner products of saliency vectors (detailed derivation can be seen in appendix). The corresponding models are noted with a suffix ``\textit{G}''. Different from the head-wise threshold, the global threshold is learnt during training and used in testing (like parameters in BN layers), thus the actual pruning rate is slightly different to the target $\xi$.
\par Table \ref{table:ImageNet-1} shows that our method outperforms all of the compared static filter pruning methods while achieving higher speed-up. For example, when compared with previous best-performed FPGM \cite{he2019filter} that achieves 1.53$\times$ speed-up, our method decreases the top-1 error by 0.37\% with 2.04$\times$ speed-up. It can also be seen that our method is superior to those dynamic execution methods including LCCN \cite{dong2017more}, FBS \cite{gao2018dynamic}, and CGNet \cite{hua2019channel}. Compared with FBS, our DGC can better achieve diversified feature representations with multiple heads that work in a self-attention way. As for LCCN and CGNet, both of them work in a pixel-level dynamic execution way (that is, the computations of some regions in the output feature maps could be skipped), which results in irregular computations and needs specific algorithms or devices for computation acceleration. However, our DGC directly removes some part of input channels for each group, which can be easily implemented like SGC. In addition, CGNet has a base path in each group that needs to be gone through by all samples, such a semi-dynamic execution mechanism may restrict its capability for capturing diversified features across different samples, while our DGC achieves fully dynamic execution through a self-attention mechanism.

\begin{figure}[t!]
\centering
    \includegraphics[width=12cm]{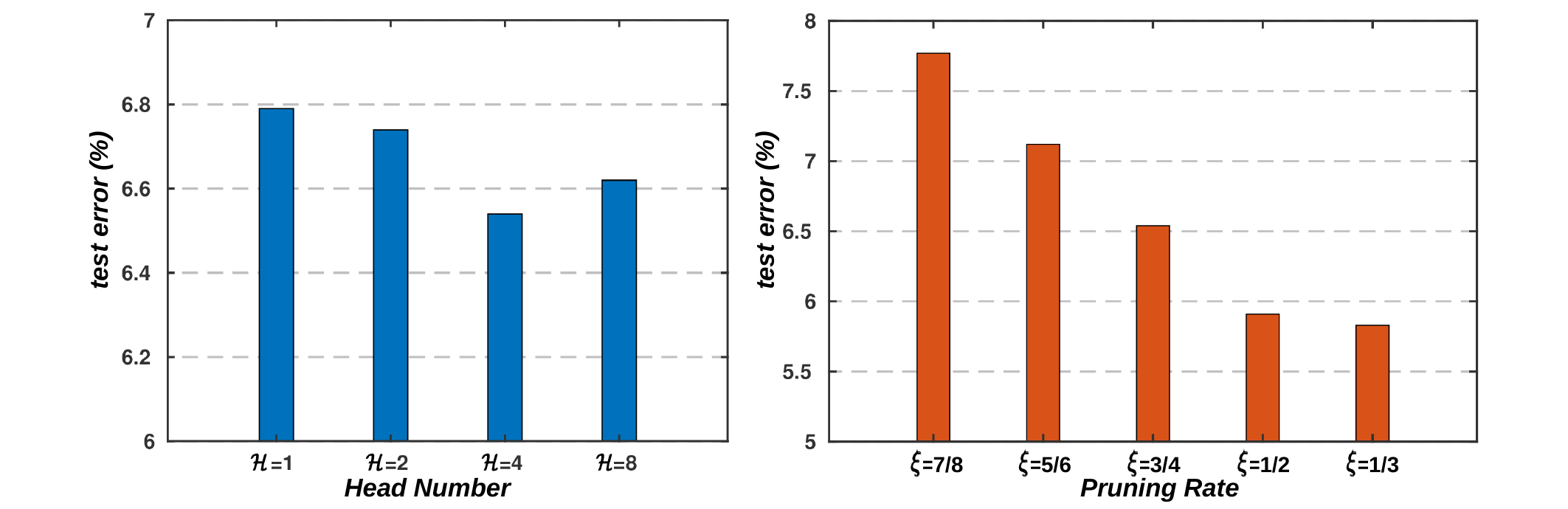}
    \vspace{-0.15in}
    \caption{Ablation study on head numbers and pruning rates}
    \label{fig:head_pr}
    \vspace{-0.15in}
\end{figure}

\begin{table}[t]
\caption{Comparison of Top-1 and Top-5 classification error with state-of-the-art efficient CNNs on ImageNet.}
\centering
\setlength{\tabcolsep}{14pt}
\resizebox{9cm}{!}{
\begin{tabular}{lccc}
\hline
\textbf{Model} & \textbf{Top-1} & \textbf{Top-5} & \textbf{MACs}\\
\hline
\hline
MobileNetV1 \cite{howard2017mobilenets} & 29.4 & 10.5 & 569M\\
ShuffleNet \cite{zhang2018shufflenet} & 29.1 & 10.2 & 524M \\
NASNet-A (N=4) \cite{zoph2018learning} & 26.0 & 8.4 & 564M \\
NASNet-B (N=4) \cite{zoph2018learning} & 27.2 & 8.7 & 488M \\
NASNet-C (N=3) \cite{zoph2018learning} & 27.5 & 9.0 & 558M \\
IGCV3-D \cite{sun2018igcv3} & 27.8 & - & 318M \\
MobileNetV2 \cite{sandler2018mobilenetv2} & 28.0 & 9.4 & 300M\\
CondenseNet \cite{huang2018condensenet} & 26.2 & 8.3 & 529M\\
CondenseNet-SGC & 29.0 & 9.9 & 529M\\
CondenseNet-FLGC \cite{wang2019fully} & 25.3 & 7.9 & 529M\\
\hline
MobileNetV2-DGC & 29.3 & 10.2 & 245M\\
CondenseNet-DGC & 25.4 & 7.8 & 549M\\
CondenseNet-DGC-G & 25.2 & 7.8 & 543M \\
\hline
\end{tabular}
}
\vspace{-0.16in}
\label{table:ImageNet-2}
\end{table}

\par It can be found from Table \ref{table:ImageNet-2} that the proposed DGC performs well when embedded into the state-of-the-art  efficient CNNs. Specifically, DGC can further speed up the MobileNetV2 by 1.22$\times$ (245M \textit{vs.} 300M) with only 1.3\% degradation of top-1 error. In addition, DGC outperforms both SGC and CondenseNet’s LGC, and obtains comparable accuracy when compared with FLGC. Finally, our MobileNetV2-DGC even achieves comparable accuracy compared with MobileNetV1 and ShuffleNet while reducing the computation cost more than doubled.

\subsection{Ablation and Visualization}
\begin{figure}[ht]
\centering
    \includegraphics[width=12cm]{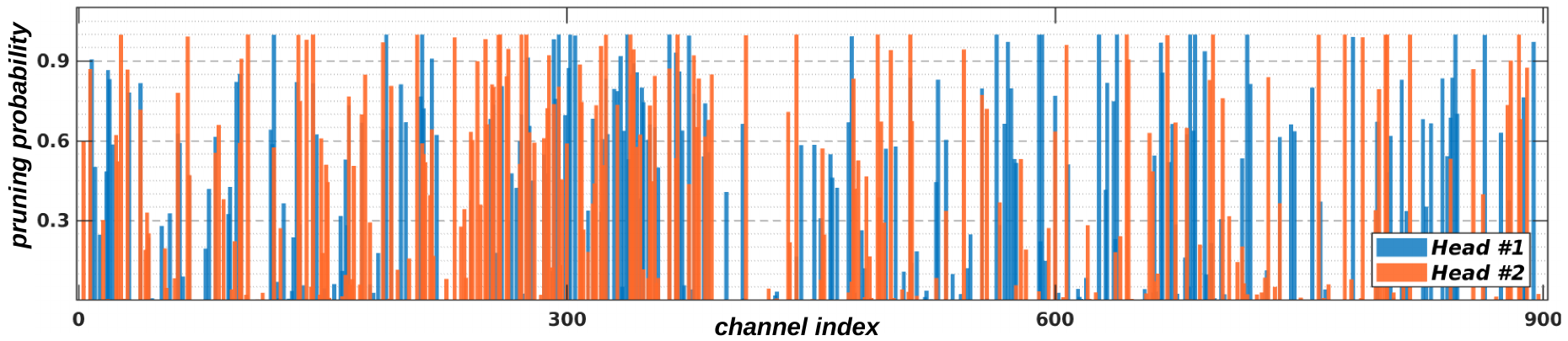}
    \vspace{-0.1in}
    \caption{Channel pruning probabilities. We track the last 900 input channels in \textit{the 33rd DGC layer} of the CondenseNet-DGC structure used in \ref{section: imagenet} on the validation set of ImageNet dataset for the first two (out of four) heads (blue and red colors). For each channel each head, the probability is calculated as the percentage of images which deactivate the channel in this head during their inferences.}
    \vspace{-0.15in}
    \label{fig:probability}
\end{figure}

\noindent{\bf Number of Heads.}
We use a 50-layer CondenseNet-DGC network structure on the CIFAR-10 dataset as the basis to observe the effects of number of heads ($\mathcal{H}=1, 2, 4, 8$). The network is trained for 150 epochs and the pruning rate $\xi$ is set to 0.75. Blue bars in Fig.~\ref{fig:head_pr} shows the Top-1 error under varying head numbers. It can be seen that the performance is firstly improved with the increase of head number, reaching the peak at $\mathcal{H}=4$, and then slightly drops when $\mathcal{H}=8$. We conjecture that under similar computation costs, using more heads helps capture diversified feature representations (also see Fig.~\ref{fig:probability}), while a big head number may also bring extra difficulty in training effectiveness. 

\noindent{\bf Pruning Rate.}
We follow the above structure but fix $\mathcal{H}=4$ and adjust the pruning rate ($\xi=7/8, 5/6, 3/4, 1/2, 1/3$). The right part in Fig.~\ref{fig:head_pr} shows the performance changes. Generally, with a higher pruning rate, more channels are deactivated during inference thus with less computation cost, but at the same time leading to an increasing error rate. 

\begin{figure}[ht]
\centering
    \includegraphics[width=12cm]{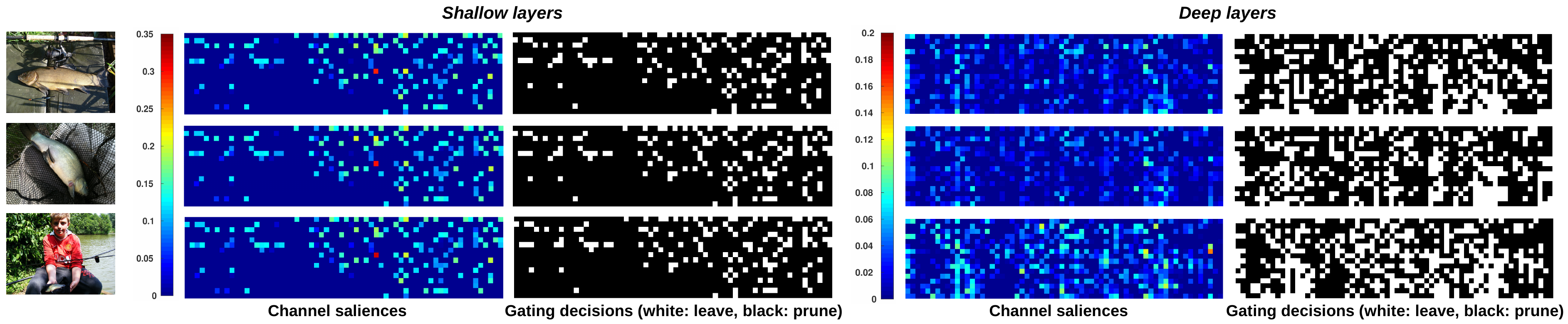}
    \vspace{-0.1in}
    \caption{Visualization of network sparse patterns for different input images. Each row represents a particular image. We compare the saliency vectors in shallow (\textit{the 5-8th DGC layers}) and deep (\textit{the 33-36th DGC layers}) DGC layers of the CondenseNet-DGC structure used in \ref{section: imagenet} and also output their corresponding pruning decisions as white-black pixels. In each saliency map (with 64 columns and 16 rows), columns mean different channels, each row indicates a certain head for a layer and every 4 rows represent a DGC layer since $\mathcal{H}=4$.}
    \label{fig:angle}
\end{figure}

\noindent{\bf \textit{Dynamicness} and \textit{Adaptiveness} of DGC networks.}
Fig.~\ref{fig:angle} visualizes the saliency vectors and corresponding sparse patterns of the CondenseNet-DGC structure used in \ref{section: imagenet} for different input images. We can see that shallower layers share the similar sparse patterns, since they tend to catch the basic and less abstract image features, while deeper layers diverge different images into different sparse patterns, matching our observation in Section~\ref{section:introduction}. On the other hand, similar images tend to produce similar patterns, and vice versa, which indicates the \textit{adaptiveness} property of DGC networks.

In addition, We also visualize the pruning probability of channels in one of the deep layers in Fig.~\ref{fig:probability}. We track the first two heads and show the \textit{jumping} channels whose pruning probability is neither 0 nor 1 (noting that those \textit{freezing} channels with probability 0 or 1 can still be jumping in the other two heads). Firstly, We can see that each head owns a particular pruning pattern but complementary with the other, indicating features can be captured from multiple perspectives with different heads. Secondly, the existence of \textit{jumping} channels reveals the \textit{dynamicness} of the DGC network, which will adaptively ignite its channels depending on certain input samples (more results can be seen in appendix).

\noindent{\bf Computation time.}
Following~\cite{huang2018condensenet,wang2019fully}, the inference speed of DGC can be made close to SGC by embedding a dynamic index layer right before each convolution that reorders the indices of input channels and filters according to the group information. Based on this, we test the speed of DGC on Resnet18 with an Intel i7-8700 CPU by calculating the average inference time spent on the convolutional layers, using the original model (11.5ms), SGC with $G=4$ (5.1ms), and DGC with $\xi=3/4$ and $\mathcal{H}=4$ (7.8ms = 5.1ms + 0.8ms + 1.9ms). Specifically, in DGC, the saliency generator takes 0.8ms and the dynamic index layer takes 1.9ms, the rest is the same as of SGC. However, the speed can be further improved with a more careful optimization.


\section{Conclusion}
In this paper, we propose a novel group convolution method named DGC, which improves the existing group convolutions that have fixed connections during inference by introducing the dynamic channel selection scheme. The proposed DGC has the following three-fold advantages over existing group convolutions and previous state-of-the-art channel/pixel-level dynaimc execution methods: 1) It conducts sparse group convolution operations while keeping the capabilities of the original CNNs; 2) It dynamically executes sample-dependent convolutions with multiple complementary heads using a self-attention based decision strategy; 3) It can be embedded into various exiting CNNs and the resulted models can be stably and easily optimized in an end-to-end manner without the need of pre-training.
\section{Appendix}
\subsection{Global Threshold and its Derivations}
\label{section:global}
This section is, for anyone's attention, a more detailed illustration about the global threshold mentioned in Section~\ref{section: GDC} and Section~\ref{section: imagenet}.

\subsubsection{Detailed derivations and Updating strategy.}
Let $\pmb{x}\in\mathbb{R}^{C\times H\times W}$, $\pmb{x}'\in\mathbb{R}^{C'\times H'\times W'}$ be the input and output of a particular DGC layer and the pruning rate is denoted as $\xi$.

For a dynamic group convolution (DGC) network, the head-wise threshold makes sure each head in the network exactly selects a certain number of channels according to the target pruning rate $\xi$ after training, \textit{i.e.,} $(1-\xi)C$ channels are selected from the input volume $\pmb{x}$ (see Eq.~\ref{eq:threshold}). While the global threshold $\mathcal{T}$ makes DGC structures more flexible allowing an uneven channel selection among heads within any DGC layer, while at the same time keeping the average pruning rate of the whole structure meeting the target $\xi$ with tiny deviation.

To obtain $\mathcal{T}$, firstly, all saliency vectors throughout the network are collected and concatenated as a single saliency vector $\pmb{G}$:
{\small \begin{equation}
    \pmb{G} = [\pmb{g}^{1,1}, \pmb{g}^{1,2}, ..., \pmb{g}^{1,\mathcal{H}}, \pmb{g}^{2,1}, ...,\pmb{g}^{\mathcal{L}, \mathcal{H}}],
\end{equation}}
where, $\mathcal{L}$ and $\mathcal{H}$ represents the number of DGC layers and number of heads in each DGC layer of the network respectively, $\pmb{g}^{i,j}$ is the saliency vector from the $jth$ head in the $ith$ DGC layer which is derived from Eq.~\ref{eq:5} but remove the ReLU activation to keep the negative saliencies for a further exploration. After that, similar to Eq.~\ref{eq:threshold}, we find the global threshold by meeting:
{\small \begin{equation}
\label{eq:global-thres}
    \xi = \frac{|\{g\;|\;\textrm{abs}(g)<\mathcal{T}, g\in\pmb{G}\}|}{|\{g\;|\;g\in\pmb{G}\}|},
\end{equation}}
where $|\mathcal{S}|$ is the length of set $\mathcal{S}$, $\textrm{abs}(z)$ returns the absolute value of $z$. 

Since saliency vectors are dynamically changing during the training process, we update the global vector every three epochs based on the last $N$ iterations at the third epoch. Therefore, assuming the batch size for each iteration is $B$, $NB$ different $\pmb{G}$s are obtained, which is regarded as the ``saliency library'' by further concatenating these $\pmb{G}$s as a new $\pmb{G}$ and put it to Eq.~\ref{eq:global-thres} to get $\mathcal{T}$. In our experiments for ImageNet, $N$ and $B$ is set as 5 and 256 respectively. This naive strategy works since the training set is randomly shuffled for each epoch, while other methods can also be tried such as introducing a running mean for $\mathcal{T}$ like the parameter updating process in batch normalization (BN) layers. Finally, like the BN layer, we adopt the finally updated $\mathcal{T}$ for inference. The experiments show that the actual pruning rate during testing is almost the same as $\xi$ (see Table~\ref{table:ImageNet-1} and Table~\ref{table:ImageNet-2}).

\subsubsection{Training with Angle Enlargements.}
We further reduce the inner product among saliency vectors from different heads within a DGC layer by introducing a third loss, in order to encourage learning saliency vectors in orthogonal directions to forcefully diversify feature representations. Our experiments show that it is automatically achieved during training if head-wise threshold is adopted, adding such loss hardly gives any improvement on the performance. However, this is slightly not the case if global threshold is applied (25.2\% \textit{vs.} 25.8\% of Top-1 error of the CondenseNet-DGC structure used in Section~\ref{section: imagenet} on ImageNet dataset with and without this angle enlargements).
Specifically, the angle enlargement loss is defined as:
\begin{equation}
    L_{a} = \lambda\frac{2}{\mathcal{L}\mathcal{H}(\mathcal{H} - 1)}\sum_{l=1}^{\mathcal{L}}\sum_{i=1}^{\mathcal{H}}\sum_{j=i+1}^{\mathcal{H}}\textrm{abs}\left(\frac{\pmb{g}^{l,i}}{\left\Vert\pmb{g}^{l,i}\right\Vert_{2}}\odot\frac{\pmb{g}^{l,j}}{\left\Vert\pmb{g}^{l,j}\right\Vert_{2}}\right),
\end{equation}
where $\odot$ represents the inner product between two vectors, $\left\Vert\pmb{z}\right\Vert_{2}$ is the $\ell^2$ norm of vector $\pmb{z}$. We set $\lambda$ to $10^{-4}$ in our experiments when using global threshold.

\subsection{More results on CIFAR datasets}

\begin{table}[t!]
\caption{Comparison of Top-1 classification error (\%) with state-of-the-art filter-level weight pruning methods.}
\centering
\setlength{\tabcolsep}{10pt}
\begin{tabular}{lccc}
\hline
\textbf{Model} & \textbf{MACs} & \textbf{CIFAR-10} & \textbf{CIFAR-100}\\
\hline
\hline
VGG-16-pruned \cite{li2016pruning} & 206M & 6.60 & 25.28\\
VGG-19-pruned \cite{liu2017learning} & 195M & 6.20 & - \\
VGG-19-pruned \cite{liu2017learning} & 250M & - & 26.52 \\
ResNet-56-pruned \cite{he2017channel} & 62M & 8.20 & - \\
ResNet-56-pruned \cite{li2016pruning} & 90M & 6.94 & - \\
ResNet-110-pruned \cite{li2016pruning} & 213M & 6.45 & -\\
ResNet-110-pruned \cite{he2019filter} & 121M & 6.15 & -\\
ResNet-164-B-pruned \cite{liu2017learning} & 124M & 5.27 & 23.91\\
DenseNet-40-pruned \cite{liu2017learning} & 190M & 5.19 & 25.28\\
DenseNet-40-pruned \cite{lin2019towards} & 183M & 5.39 & -\\
DenseNet-40-pruned \cite{lin2019towards} & 81M & 6.77 & -\\
CondenseNet-86 \cite{huang2018condensenet} & 65M & 5.00 & 23.64\\
\hline
CondenseNet-86-DGC & 71M & \textbf{4.77} & \textbf{23.41}\\
CondenseNet-86-DGC-G & 71M & \textbf{4.42} & \textbf{23.36} \\
\hline
\end{tabular}
\label{table:condensenet86}
\end{table}

We further evaluate our method on the CondenseNet-86 structure used in \cite{huang2018condensenet} by replacing the learnt group convolution (LGC) of CondenseNet with DGC, and compare it with the original CondenseNet and other state-of-the-art filter-level pruning methods. The parameter settings are the same as the CondenseNet. Results are shown in Table~\ref{table:condensenet86}. In this table, the model with a suffix ``G'' means we adopt the global threshold. 

\subsection{Further Visualization}
\begin{figure}[t!]
    \includegraphics[width=12cm]{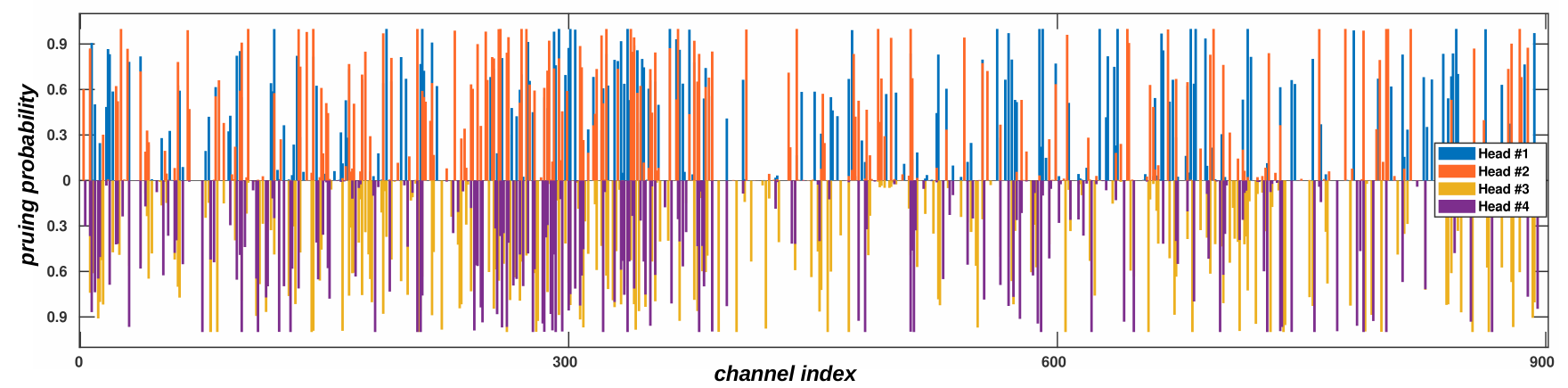}
    \caption{Extension for Fig.~\ref{fig:probability} in the original paper, with the other two heads visualized.}
    \label{fig:prob}
\end{figure}

\begin{figure}[t!]
    \includegraphics[width=12cm]{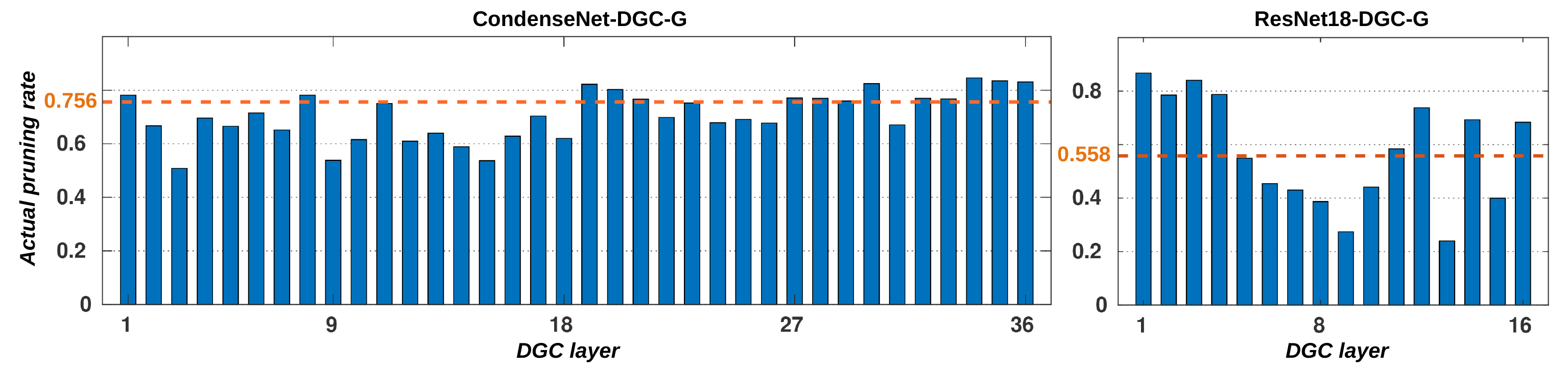}
    \caption{Actual pruning rate of each DGC layer for CondenseNet (left) and ResNet (right) structure on the validation set of ImageNet dataset using the global threshold. \textit{CondenseNet-DGC} corresponds to the model in Table~\ref{table:ImageNet-2} with the same name. \textit{ResNet18-DGC-G} corresponds to the model \textit{DGC-G} in Table~\ref{table:ImageNet-1} from the original paper. The red line represents the overall pruning rate for the model. It can be seen that by using global threshold, the network is given more flexibility that allows each layer adapting to a particular pruning rate, leading to a slightly better performance than the one with head-wise thresholds, but at the same time bringing extra irregularity to the model structure (\textit{e.g.}, even within a single layer, different input smaples may also lead to different numbers of channels selected). A balance between such irregularity and performance need to be considered during network design.}
    \label{fig:pruningrate}
\end{figure}


\bibliographystyle{splncs04}
\bibliography{DGC_arXiv}
\end{document}